# Machine Learning-Based Intrusion Detection and Prevention System for IIoT Smart Metering Networks: Challenges and Solutions

Sahar Lazim, Qutaiba I. Ali
University of Mosul
Iraq

**Abstract:**
The Industrial Internet of Things (IIoT) has revolutionized industries by enabling automation, real-time data exchange, and smart decision-making. However, its increased connectivity introduces cybersecurity threats, particularly in smart metering networks, which play a crucial role in monitoring and optimizing energy consumption. This paper explores the challenges associated with securing IIoT-based smart metering networks and proposes a Machine Learning (ML)-based Intrusion Detection and Prevention System (IDPS) for safeguarding edge devices. The study reviews various intrusion detection approaches, highlighting the strengths and limitations of both signature-based and anomaly-based detection techniques. The findings suggest that integrating ML-driven IDPS in IIoT smart metering environments enhances security, efficiency, and resilience against evolving cyber threats.

## 1. Introduction

Everything globally, from body sensors to contemporary cloud computing, is included in the Internet of Things (IoT). It creates a sophisticated distributed system by connecting humans, machines, and networks everywhere; it improves the quality of human life by enabling reliable machine-to-machine and machine-to-human connections [1]. The integration of conventional Internet of Things (IoT) principles in manufacturing industries and applications is referred to as the Industrial Internet of Things (IIoT) [2]. In the recent decade, the IIoT has emerged as the most rapidly evolving revolutionary technology, with the ability to digitize and connect numerous industries, resulting in significant economic prospects and global Gross Domestic Product (GDP) growth [3, 4]. Smart metering, smart factories, smart grids, smart cities, smart homes, linked autos, and supply chain management are just a few of the IIoT's applications [2].

The vast number of sensors in the IIoT network generates a tremendous volume of data [3]. The edge devices collect data from appliances and send it to local servers after performing any essential preprocessing. Then, local severs are utilized to connect edge devices and cloud servers. [2].

With the broad deployment of smart meters, it is now possible to add the smart meter's function to the edge device's function. The smart meter is one of the main components of the smart metering network, which is used to learn regular power operating usage and utilize it to detect or flag abnormal usage. User behavior, human mistakes, and underperforming equipment contribute to wasted energy in buildings and industries. Identifying anomalous consumption power behavior can help reduce peak energy usage and change undesirable user behavior. The smart meter data's average power consumption characteristic distribution is relatively regular and has noticeable periodicity, but aberrant power consumption lacks these features. Therefore, it can be assumed that often-observed daily patterns represent usual consumption behaviors during the day, while rarely-observed patterns represent unusual consumption behaviors [5].

Monitoring and regulating such infrastructures simultaneously is important to provide an adequate degree of security for IIoT networks. Intrusion Detection System and Intrusion Prevention System (IDS/IPS) are the most often used systems for achieving this goal [6]. Where IDS monitors intrusions and occurrences of safety violations and notifies people when they happen. On the other hand, IPS takes further steps to avoid an attack, mitigate its consequences, or respond actively to a security breach [7, 8].

Signature-based detection and anomaly-based detection are the two common detection methods used by IDS. By evaluating network data for specified patterns, signature-based detection approaches are successful in detecting well-known threats. Attacks are detected via anomaly-based detection systems, which monitor the behavior of the entire system, traffic, or objects and compare it to a preset normal state [9-11]. Anomaly-based IDS provides greater generic properties than signature-based IDS because of Machine Learning (ML) approaches [10, 12].

An important direction is the use of ML algorithms for anomaly-based IDS approaches. The general aspects of the training traffic information are learned by ML techniques. The input traffic information will be accurately detected based on the learned attributes. Conventional ML methods, on the other hand, do not demand a lot of processing power from the hardware and have a short training time, making them more suited to the needs of IIoT edge devices [13, 14].

However, an intelligent, robust, secure, and resilient Intrusion Detection and Prevention System (IDPS) is required for IIoT edge layer devices.

The objectives of this paper are:
- To analyze security threats and vulnerabilities in IIoT-based smart metering networks.
- To review existing intrusion detection methodologies and their applicability in IIoT environments.
- To evaluate the effectiveness of the proposed system through performance metrics such as detection accuracy, computational efficiency, and response time.

## 2. IoT Application Domain

IoT applications will evolve continuously over time but they must be faced numerous challenges relevant to security, privacy, complexity, scale, spectrum enough to connect a large number of sensors or tagged objects, etc.[15]. The application evolved for IoT must be able to handle the data in real-time and connect with the other devices. Moreover, it must not only treat with actuating and sensing but rather it will cooperate with the human also, for example, Human to Machine (H2M), Human to Human (H2H), and Machine to Machine (M2M) [16]. The domain of IoT application is not only limited to one side of human life such as the concept of building as a smart home or smart building but also prevalent in other domains such as agriculture, health, smart businesses, banking, environmental monitoring, etc. [16].

The IoT applications can be categorized based on the network availability type, scale, coverage, repeatability, heterogeneity, user impact, and involvement. Thus, IoT applications are categorized into four application fields: personal and home, enterprise, utilities, and mobile. Figure1 shows that the personal and home class represented IoT at the scale of an individual or home, enterprise IoT at the scale of a community, utility IoT at a national or regional scale, and mobile IoT that is usually spread through other classes fundamentally because of the scale and connectivity nature [17, 18].

IoT applications offer several capabilities and functionalities that are grouped depending on the utilization domain into four fields: monitoring (environment state, devices condition, alert, notifications, etc.), control (devices functions control), optimization (device diagnostics, repair, performances, etc.) and autonomy (autonomous operations) [19, 20].

Table1 conducts the major IoT application domains presently specified by the research community along with some examples for each field [21, 22]. There is a number of use-cases for each application domain that have been implemented. Furthermore, many planned use-cases will be implemented in the coming years [17].

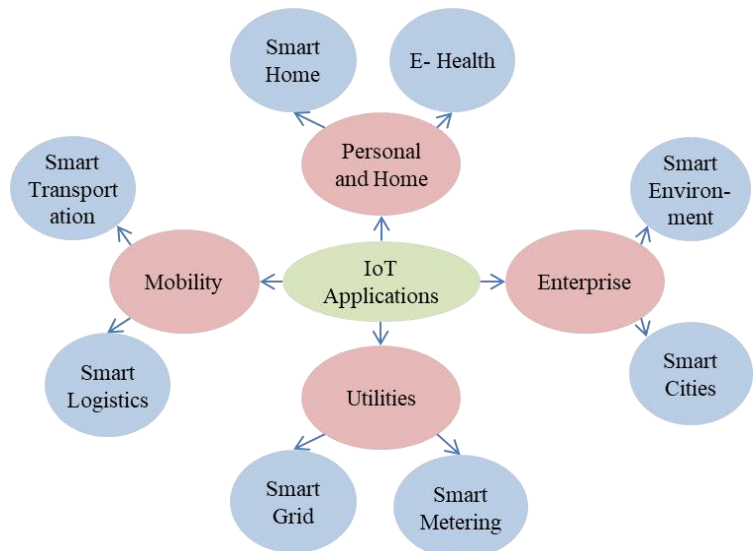

Figure1: IoT application domain and their examples

Smart Metering Network is a part of electrical power communications that integrates power and telecom infra to collect, monitor, and evaluate customer power usage. It includes smart meters, a head-end system, and data concentrators and is organized into numerous communications networks [23, 24]. It is composed of a Home Area Network (HAN), which interconnects appliances with smart meters, Neighborhood Area Network (NAN), which represents the network interconnecting the smart meters with the data concentrator, and Wide Area

Networks (WAN) interconnects multiple NANs to the Utility headend [25]. A wired or wireless connection is used for bidirectional communication between the devices. The WAN is primarily based on long-distance and high-speed communication in terms of communication rate, and the data sent which is sensitive information.

Table1 The main IoT application domains.

| Application Domain | Examples of Application |
| --- | --- |
| Home Automation | Remote control applications, Building automation, Water use, and Energy use |
| Retail & Logistics | Stock control, Smart shopping, E-Payments, Supply chain control, Fleet tracking, Item tracking, and Smart product management. |
| E-Health | Patient monitoring, Patient surveillance, Personnel tracking, Doctor tracking, Predictive expertise information to assist doctors and practitioners, and Real-time patient health status monitoring [26]. |
| Smart Animal Farming | Meat traceability and Animal monitoring |
| Security and Emergency | Perimeter control and tracking, and Disaster recovery |
| Smart Cities | Structural health, Traffic congestion, Participatory sensing, Smart traffic lights, Smart parking management, and Accident detection |
| Smart Transportation | Smart transportation through real-time dynamic on-demand traffic information and shortest-time travel path optimization. |
| Smart Metering | Metering of heating systems, and Metering in the smart electric grid |
| Industrial Control | Mobile robotics applications in industry, and Manufacturing applications |
| Smart Environment | Forest fire detection, Remote seismography, Pollution monitoring, Air pollution, and Noise monitoring, |
| Smart Water | Level-monitoring, Flood detection, and Water leakages detection |
| Smart Agriculture | Moisture management, Crop monitoring, Compost, Soil, Irrigation management, and Smart greenhouses. |

The NAN uses Short-distance and low-speed communication. Also, the HAN is primarily low-speed, short-distance communication. As the HAN's intelligent appliances are connected to the Internet, they are more susceptible to DoS and other threats. The HAN uploads its internal data collection and has limited computing and storage capacity [27]. Figure2 displays the general structure of the smart metering network.

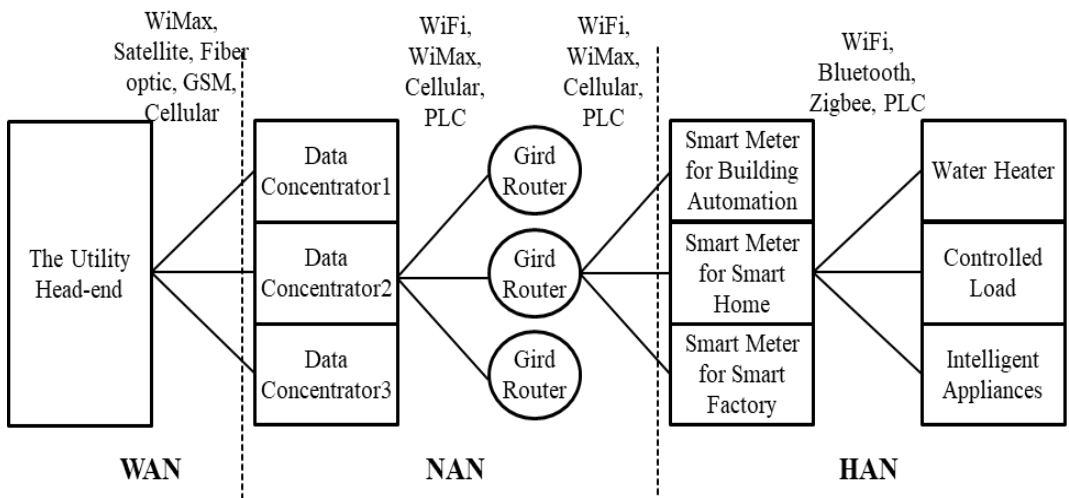

Figure 2: Communication structure of smart metering network

## 3. Literature Review

Various studies have yielded many approaches for IDSs in the security of IoT/IIoT networks. So, this paper will focus on three domains (IoT, IIoT, and Smart Metering). The relationship between these domains is shown in Figure 3. As a result, this section is divided into three parts based on Figure4 that shows the numbers of subsections.

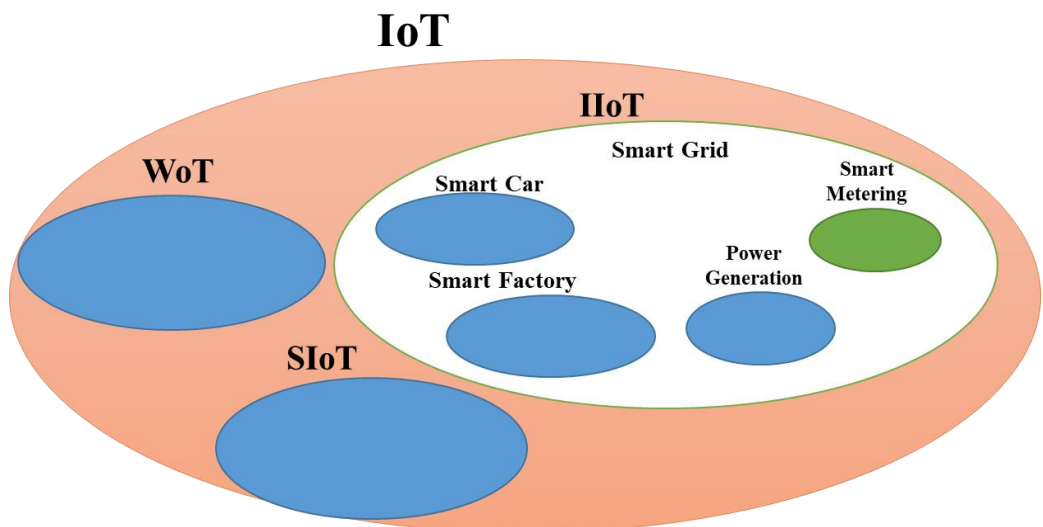

Figure 3: The relationship between (IoT, IIoT, and smart metering)

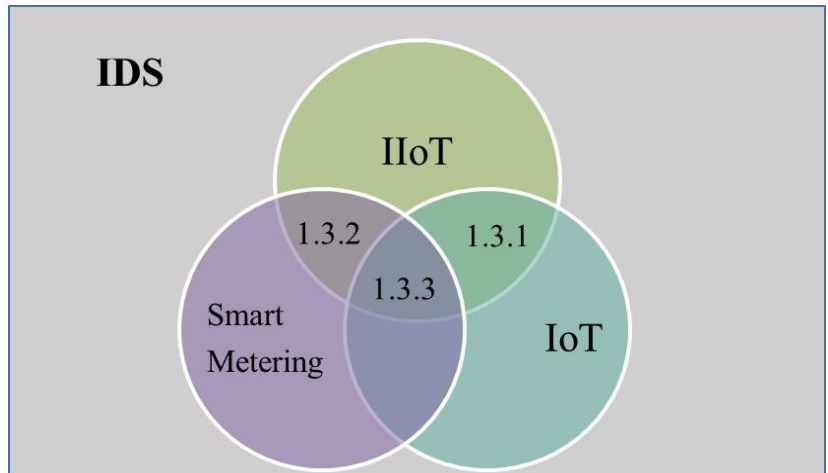

Figure 4: The multiple IDS approaches for the literature review

**3.1 IDS based IoT/IIoT Networks**

In recent years, several researchers have proved that ML models were used to construct IDSs to safeguard IoT/IIoT networks [28, 29]. A brief survey of the literature is offered here:

(Aydogan, et al., 2019) [30] proposed a genetic programming-based intrusion detection methodology for IIoT systems, as well as a proof of concept and quantitative assessment. Furthermore, because the network involves a variety of devices with varying capabilities and resources limits, the appropriateness of a central IDS at the root node is investigated. It was demonstrated that the root node could efficiently identify RPL attacks in a timely way. (Shalaginov, et al., 2019) [31] developed a new method that allows for the benefits of ML techniques for IoT security while balancing the performance of the model and computational complexity. They proposed the use of the Message Queue Telemetry Transport (MQTT) as the basis for a protocol for updating neural network models on resource-constrained devices. This allows all unessential and demanding model learning and meta-heuristic improvement operations to be moved to more capable IoT gateway devices. Simultaneously, the IoT node devices will undergo the less computationally intensive testing phase.

(Latif, et al., 2020) [32] proposed a unique IIoT attack prediction system based on a lightweight random neural network. Using the DS2OS dataset, the suggested system identifies IIoT threats with high accuracy and reduced forecast time. The suggested algorithm's performance is assessed by generating several performance parameters with varied limitations. (Maharani, et al., 2020) [33] suggested identifying attacks in a fog computing environment using numerous ML methods to efficiently detect harmful activity. The evaluation was carried out using the KDD Cup'99 dataset, with K-Means, Random Forest (RF), and Decision Tree (DT) algorithms being compared. According to (Eskandari, et al., 2020) [34], Passban is an intelligent IDS that can secure IoT devices that are directly linked to it. The suggested method is unique in that it may be installed directly on low-cost IoT gateways (e.g., single-board PCs). Passban can identify a variety of malicious activity, such as SSH (Secure Shell) and HTTP (Hypertext Transfer Protocol) Brute Force, TCP Synchronous (SYN) Flood , and Port Scanning assaults, with a low false positive rate and high accuracy.

(Awotunde, et al., 2021) [35] developed a deep learning-based attack detection framework for IIoT networks. A combination of rule-based feature engineering and a deep feed-forward neural network model was used to implement the training procedure. NSL-KDD and UNSW-NB15 datasets were used to evaluate the suggested approach. (Raja, et al., 2021) [36] introduced a two-stage deep learning-based IDS for IIoT networks. Deep Neural Networks (DNN) are trained and evaluated in the first stage of detection. Second-stage detection is fed assaults with a reduced detection rate or precision in first-stage detection. The DNN and Negative Selection Algorithm are trained utilizing Dragonfly Algorithm and are employed for second-stage detection. The outputs of both algorithms are combined using Dempster Shafer's combination rule. They compared their findings to datasets from CICIDS 2017, CICIDS 2018, and TON IoT.

(Nguyen, et al., 2022) [37] presented Realguard IDS, a DNN-based network IDS directly run on local gateways to safeguard IoT devices within the network. With a small processing footprint, their concept can accurately identify several cyber-attacks in real-time. This is accomplished using a simple feature extraction approach and a DNN-based assault detection model. Realguard could identify 10 types of attacks (e.g., port scan, Botnet, and FTP-Patator) in real-time and operate on resource-constrained gateways (Raspberry Pi), according to their analyses on the CICIDS2017 dataset.

Based on Table2, numerous researchers did a fine job presenting effective intrusion detection strategies for IoT/IIoT platforms. Nevertheless, there are certain limits to the existing researches. First, most researchers in recent studies evaluated their schemes by achieving a few performance metrics that do not offer in-depth evaluation. Second, the presented techniques' feasibility for resource-limited devices is not thoroughly deliberated. Finally, several studies' great experimental accuracy is due to a mixture of feature engineering techniques such as feature mapping and feature reduction, with classification algorithms. On the other hand, feature engineering is a time-consuming process that is not suitable for time-critical edge layer devices that have imperfect computational capabilities.

Table 2 The details for the recent research based on IDS for IoT/IIoT networks

| Ref. | Year | Dataset Used | Algorithm Utilized | Updating Policy | Architecture Approach | Performance Parameters | Prevention and Response activities | Validation Strategy | Performance Metrics |
|---|---|---|---|---|---|---|---|---|---|
| [30] | 2019 | N/A | Genetic Programming | No | Centralized | Detection Efficiency | No | No | Acc=99.83% |
| [31] | 2019 | NSL-KDD | Neural Network | Yes | Distributed | N/A | No | Yes | N/A |
| [32] | 2020 | DS2OS | Random Neural Network | No | Centralized | Detection Efficiency, Resource Usage | No | Yes | Acc=99.20% |
| [33] | 2020 | KDD Cup'99 | K-Mean | No | Centralized | Detection Efficiency | No | No | Acc=93.33% |
| [34] | 2020 | Generated by author | Isolation Forest and Local Outlier Factor | Yes | Distributed | Detection Efficiency, Resource Usage | No | Yes | N/A |
| [35] | 2021 | NSL-KDD and UNSW-NB15 | DNN | No | Centralized | Detection Efficiency | No | No | Acc1=99% Acc2=98.9% |
| [36] | 2021 | CICIDS 2017, CICIDS 2018, TON IoT | DNN | No | Centralized | Detection Efficiency | No | No | Acc1=99.86% Acc2=97.58% Acc3=98.8% |
| [37] | 2022 | CICIDS2017 | DNN | Yes | Distributed | Detection Efficiency, Resource Usage | Yes | Yes | Acc=99.64% |

N/A: Not Available.

## 3.2 IDS for IIoT Networks with Smart Metering

A growing number of researchers have used machine learning and data mining approaches to model and evaluate household power consumption data during the last few years [38]. Consequently, several research projects have been displayed in this section.

(Jokar, et al., 2016) [39] suggested a new Consumption Pattern-Based Energy Theft Detector (CPBETD), which relied on the consistency of normal users as well as an example of a malicious consumption pattern. The test data-target value was predicted using the Support Vector Machine (SVM) approach. The dataset for this study comes from the Irish Smart Energy Trial. The total performance of the CPBETD system employed in this study has high accuracy and high detection rate.

(Bhattacharjee, et al., 2017) [40] identified four forms of data falsification threats: conflict, camouflage, deductive, and additive. Pecan Street dataset was utilized in their investigation to detect data falsification threats using statistical techniques. Their final result effectively reveals a high detection rate, with low false positives.

(García, et al., 2018) [41] experimentally investigated the effect of unsupervised and supervised neural network models to identify unexpected electricity consumption habits. Furthermore, investigations show that the supervised technique significantly improves the anomaly identification rate. Identifying consumption abnormalities was highlighted as a real-time extensive data analysis issue by (Xiang, et al., 2019) [42]. Furthermore, utilizing the in-memory distributed computing framework Spark and its extension Spark Streaming, the author developed a supervised learning and statistical-based anomaly detection method and implemented a lambda system. (Chahla, et al., 2019) [43] created an anomaly detection method to increase energy efficiency and detect aberrant activities. The K-means method trains numerous situations that describe each user's energy consumption pattern. The Long Short Term Memory (LSTM) algorithm is utilized to anticipate the next hour's power consumption. Real-world power consumption data from Pecan Street in the United States is used to test the concept.

(Himeur, et al., 2020) [44] proposed a new method for detecting energy usage anomalies based on using a rule-based model to extract micro-moment characteristics. Leveraging daily intent-driven moments of user-consuming activities extracts output characteristics. They experiment with a deep learning model for effective abnormal identification and classification, in addition to micro-moment extraction of features.

(Oprea, et al., 2021) [45] suggested two unsupervised machine learning algorithms methods for detecting anomalies in unlabeled data: the Spectral Residual-Convolutional Neural Network (SR-CNN) and a martingales-based anomaly trained model for determining changes in time-series streaming data. The Fisher Linear Discriminant Analysis and the Two-Class Boosted Decision Tree are deployed to the recently processed dataset. (Zhang, et al., 2021) [46] integrate the well-known deep learning model transformer with the clustering technique K-means to predict energy usage over time and identify abnormalities. The transformer model is utilized to predict energy consumption for the next hour, and the K-means clustering algorithm is employed to improve forecasting results. Lastly, anomalies are determined by analyzing predicted and tested values.

Based on Table 3, using aggregated-level data to detect energy usage aberrations is not the preferred approach since it is widespread and does not provide accurate information about the causes of each irregularity. As a result, using information collected at the appliance level by smart meters is desirable since it helps in the detection of anomalies in each device. Furthermore, there have been no platforms for generating empirical data, which is one of the most difficult aspects of power consumption anomaly detection.

Table 3 Comparison frameworks conducted in anomaly power consumption detection.

| Ref. | Year | Dataset Used | Algorithm Utilized | Updating Police | Network Field | Place on | Anomaly Type | Validation Strategy | Performance Metrics |
|---|---|---|---|---|---|---|---|---|---|
| [39] | 2016 | Irish Smart Energy Trial | Multiclass-SVM | No | NAN | Data Aggregator | Malicious Consumption Patterns | No | DR=94% |
| [40] | 2017 | Pecan Street | Statistical Technique | No | N/A | N/A | Data Falsification | No | N/A |
| [41] | 2018 | UC Irvine | Neural Network | No | N/A | N/A | Unexpected Electricity Consumption | No | N/A |
| [42] | 2019 | EERE | Clustering Technique | No | NAN | Data Aggregator | Abnormal Electricity Consumption | No | Acc=99.6% |
| [43] | 2019 | Pecan Street | K-Means and LSTM | No | NAN | N/A | Aberrant Activities | No | N/A |
| [44] | 2020 | QUD, DRED, and PCSiD | Deep Learning | No | NAN | Data Aggregator | Energy Consumption Anomalies | No | Acc=99.58% |
| [45] | 2021 | CER and ISSDA | SR-CNN | No | NAN | N/A | Energy Consumption Anomalies | No | Acc=90% |
| [46] | 2021 | UC Irvine | K-Means and Deep Learning | No | NAN | N/A | Energy Consumption Anomalies | No | Acc=97% |

QUD: Qatar University dataset  
EERE: Office of Energy Efficiency & Renewable Energy  
UCI: UC Irvine machine learning repository  
CER and ISSDA : Commission of Energy Regulation and Irish Social Science Data Archive

### 3.3 IDS in IoT/IIoT Networks for Smart Metering:

Many recent works have noticed the security problem of IoT/IIoT networks for smart metering application, and many intrusion detection methods have been proposed and developed [47].

(Sedjelmaci, et al., 2016) [48] presented a new method for detecting the attacks in the smart grid Neighborhood Area Network (NAN). To protect the communications and power components in grid NAN, the proposed detection technique operates at control centers, smart meters, and collectors. For training and classification, a mix of rule-based detection and a learning algorithm is used based on the KDD99 dataset. The technique had a high protection level and required little energy in the simulation. (Jindal, et al., 2016) [49] presented an energy theft detection system using a SVM and a DT. At all energy transmission and distribution stages, the fusion DT-SVM classifier is equipped to acceptably identify real-time power theft.

A Zigbee-based intrusion detection and prevention model for Home Area Networks (HANs) in the smart metering network was suggested by (Jokar, et al., 2016) [50]. The model employs a Q-learning-based intrusion detection system that learns the best approach for dealing with intruders through interactions with the environs to secure the network from intruders.

(Vijayanand, et al., 2017) [51] utilized an IDS built with a multi-Support Vector Machine (SVM) classifier to identify attacks in the smart metering network. Each classifier recognizes a single type of attack. The improved intrusion detection system's performance is assessed through training and testing states the classifier by the ADFA-LD dataset. According to simulation results, the suggested intrusion detection technique is highly suited for reliably identifying attacks in NAN of smart metering.

(Li, et al., 2018) [52] presented a model for detecting Advance Metering Infrastructure (AMI) intrusions based on the Online Sequence Extreme Learning Machine (OS-ELM). The model utilized online sequence training to improve detection speed while maintaining accuracy.

(Vijayanand, et al., 2019) [53] developed a unique IDS with multiple deep learning layers structured in hierarchical sequence and a high attack detection ratio for each type of attack in smart metering network traffic. CICIDS 2017 dataset is used to assess the performance of the proposed system.

By integrating the Convolutional Neural Networks (CNN) with the Gated Recursive Unit (GRU) and the Particle Swarm Optimization (PSO) method, a Hybrid Deep Neural Network (HDNN) intrusion detection model is created by (Ullah, et al., 2020) [54]. The CNN selects and extracts features, while the GRU-PSO technique categorizes the given data.

A new IDS for smart meters is presented by (Sun, et al., 2020) [55], where it comprises two detection stages for detecting malicious activities. The SVM approach is utilized in the initial step to detect suspicious activities in smart meters and then report them to the second stage. The IDS algorithm stage computed the matching between observed anomalous activities and pre-defined intrusion actions. The test platform includes Network Simulator 3 program, which simulates smart metering network traffic. At the same time, a Single-Board Computer (SBC) is used to mimic the IEEE 802.15.4 connection between a smart meter and a grid router.

(Yao, et al., 2021) [27] suggested a Convolutional Neural Networks and Long Short-Term Memory Intrusion Detection (CNN-LSTM ID) model with cross-layer feature-fusion. The experiments show that the suggested cross-layer feature-fusion CNN-LSTM ID model outperforms standard ID models when tested on two datasets, the NSL-KDD and KDD Cup 99.

Based on Table 4, the majority of the researchers presented their attack detection techniques by focusing on either specific network attacks or anomalous power consumption behavior rather than both. Furthermore, they primarily used old publicly available datasets such as KDD Cup 99, NSL-KDD, and others to evaluate their models. As a result, new datasets are necessary to meet the current security requirements of smart metering networks. Another limitation of the mentioned related works is that most researchers did not discuss how their proposed models would work with resource-constrained devices. At the same time, the anomaly detection technique should consider the limited processing resources for smart meters. Even though existing IDSs in the related works provide good protection capabilities against cyber threats and abnormal power consumption behavior, the detection function may consume substantial computational resources and impact smart meter operation.

Table 4 The details for the recent researches based on IDS for smart metering network

| Ref. | Year | Dataset Used | Algorithm Utilized | Network field | Placed on | Attacks Detected | Validation Strategy | Performance Metrics |
|---|---|---|---|---|---|---|---|---|
| [48] | 2016 | KDD99 | SVM and Rule-based | NAN | Smart meter, data aggregation, and head-ends | DoS and Probe | No | Variant for each level |
| [49] | 2016 | Custom | Fusion DT-SVM | NAN | Data aggregator | Abnormal Behaviors | No | Acc=92.5% |
| [50] | 2016 | Custom | Q-learning | HAN | Smart meter | Spoofing, physical Eavesdropping DoS | No | N/A |
| [51] | 2017 | ADFA-LD | multi-SVM | NAN | Data aggregator | Exploits, DoS, Fuzzers, Backdoor Generic, Worms | No | Acc=90% |
| [52] | 2018 | ISSDA | OS-ELM | NAN | Data aggregator | Abnormal Behaviors | No | Acc=98.75% |
| [53] | 2019 | CICIDS 2017 | deep learning and SVM | NAN | Data aggregator | DoS, PortScan, Web Attack, Bot, FTP Parator, SSH Parator | No | Acc=99.99% |
| [54] | 2020 | SGCC | HDNN | NAN | Data aggregator | Electricity Theft Detection | No | Acc=89% |
| [55] | 2020 | SCT | SVM | NAN | Smart meter | Abnormal Behaviors | Yes | Acc=98.71% |
| [27] | 2021 | KDD Cup 99 and NSL-KDD | CNN-LSTM | WAN | Head-ends | DoS, probe, R2L, U2R | No | Acc1=99.95% Acc2=99.79% |

## 4. Analysis of Review Findings

The review of existing studies on Intrusion Detection and Prevention Systems (IDPS) in IIoT-based smart metering networks reveals key insights into the current state of cybersecurity in industrial environments. This section summarizes the findings, highlights limitations, and identifies research gaps that must be addressed to develop an effective and scalable security solution.

### 4.1 Effectiveness of Machine Learning in Intrusion Detection

Machine Learning (ML) techniques have been widely adopted for intrusion detection due to their ability to recognize patterns and detect anomalies beyond traditional rule-based methods. Studies demonstrate that anomaly-based detection using ML models provides greater adaptability to emerging cyber threats than signature-based methods, which rely on predefined attack patterns. However, several challenges persist [56]:

- Many ML-based IDS solutions require high computational power, limiting their deployment on resource-constrained edge devices.
- The lack of standardized datasets for IIoT smart metering security hinders the generalizability of ML models.
- While deep learning models have demonstrated high accuracy, they often require extensive training and large-scale data, which may not always be feasible in real-time IIoT environments.

### 4.2 Security Challenges in Smart Metering Networks

Smart metering networks form a critical component of IIoT infrastructure, yet they face multiple security threats, including [57]:

- Cyberattacks such as Denial of Service (DoS), data falsification, and unauthorized access.
- Data privacy concerns due to the continuous transmission of consumer energy usage patterns.
- Scalability issues where increasing the number of smart meters can strain security mechanisms and cause delays in threat detection.
- Limited computational resources in edge devices that restrict the implementation of complex IDS algorithms.

### 4.3 Existing Approaches and Their Limitations

A comparative review of various intrusion detection methods reveals several trends [58]:

- Signature-based IDS provides quick and accurate detection for known attacks but fails to recognize novel threats.
- Anomaly-based IDS using ML techniques offers higher adaptability but often suffers from higher false positive rates.
- Hybrid IDS approaches combining rule-based and ML-based detection methods show promising results but require further optimization to balance accuracy and computational efficiency.
- Few studies address real-time detection in smart metering networks, focusing primarily on offline data analysis rather than live threat mitigation.

### 4.4 Need for Adaptive and Lightweight IDPS Solutions

To address these challenges, a next-generation **lightweight and adaptive IDPS** is required, integrating the following features [59]:

- Resource-efficient ML models that can operate on edge devices with limited power and memory.
- Continuous model updates using online learning techniques to enhance detection capabilities over time.
- Federated learning approaches to enable collaborative anomaly detection across distributed smart metering networks without compromising data privacy.
- Integration with blockchain or distributed ledger technologies to ensure the integrity and immutability of security logs.

## 5. Research Gaps and Future Directions

The review findings suggest that while significant progress has been made in securing IIoT-based smart metering systems, the summary of key finding is shown in Table 5, and the following research gaps need further exploration [60]:

- **Developing real-time, low-latency intrusion detection mechanisms** that can efficiently detect and mitigate threats with minimal impact on system performance.
- **Creating benchmark datasets for IIoT smart metering security** to enhance model training and evaluation consistency.
- **Exploring energy-efficient deep learning architectures** that reduce computational overhead while maintaining high detection accuracy.
- **Investigating cross-domain applicability** by extending IDPS solutions to other critical IIoT applications, such as smart grids and industrial automation.

Table 5: Summary of Key Findings

| Category | Current Trends | Challenges & Gaps |
|---|---|---|
| Intrusion Detection | ML-based anomaly detection | High false positives, resource constraints |
| Smart Meter Security | Data-driven monitoring | Privacy risks, cyberattack vulnerabilities |
| Computational Efficiency | Lightweight ML models | Limited processing power on edge devices |
| Scalability & Adaptability | Hybrid IDS approaches | Need for real-time and adaptive learning |
| Future Security Enhancements | Federated learning, blockchain | Lack of large-scale deployment in IIoT |

## 6. Conclusions

The analysis highlights the growing need for robust, adaptive, and resource-efficient IDPS solutions tailored for IIoT-based smart metering networks. While ML techniques provide promising results in intrusion detection, further advancements are required to enhance real-time performance, scalability, and energy efficiency. Addressing these research gaps will enable the development of a next-generation cybersecurity framework capable of protecting IIoT infrastructure against evolving threats.